\documentclass[10pt,twocolumn,letterpaper]{article}

\usepackage{cvpr}              

\usepackage{graphicx}
\usepackage{amsmath}
\usepackage{amssymb}
\usepackage{booktabs}
\usepackage{subcaption}
\usepackage{comment}

\usepackage[pagebackref,breaklinks,colorlinks]{hyperref}

\usepackage[capitalize]{cleveref}
\crefname{section}{Sec.}{Secs.}
\Crefname{section}{Section}{Sections}
\Crefname{table}{Table}{Tables}
\crefname{table}{Tab.}{Tabs.}

\def\cvprPaperID{0036} 
\def\confName{CVPR}
\def\confYear{2023}

\begin{document}

\title{CustomText: Customized Textual Image Generation using Diffusion Models}

\author{Shubham Paliwal, Arushi Jain, Monika Sharma, Vikram Jamwal, Lovekesh Vig\\
TCS Research\\
New Delhi, India\\
{\tt\small (shubham.p3, j.arushi, monika.sharma1, vikram.jamwal, lovekesh.vig)@tcs.com}
}

\twocolumn[{%
\renewcommand\twocolumn[1][]{#1}%
\maketitle
\begin{center}
    \centering
    \captionsetup{type=figure}
    \includegraphics[width=0.95\textwidth]{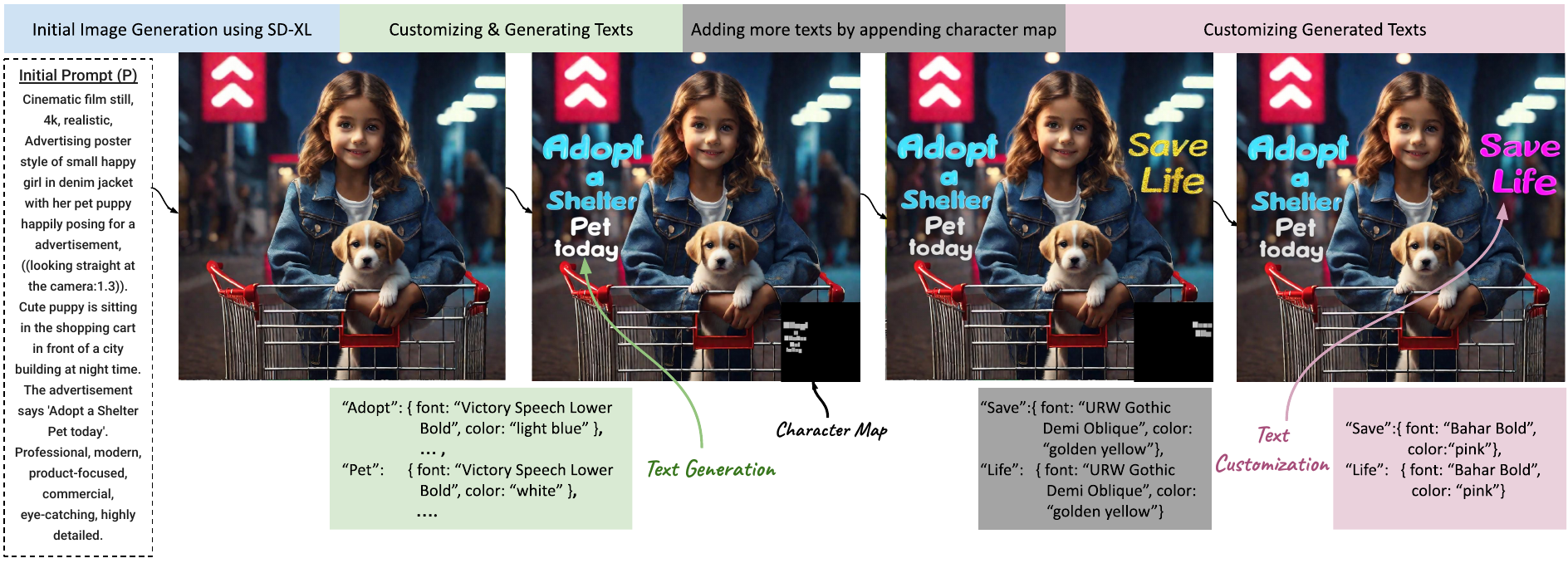}
    \vspace{-2mm}
    \captionof{figure}{\small{An example illustrating the use of our proposed CustomText method. It is a simulation of an advertisement designing workflow for an Ad campaign ``Adopt a Pet", where initial image (A) is generated using text-prompt, P by stable diffusion base model~\cite{sdxl}. Subsequently, the user can customize the font attributes of the text by using a user-interface (B). User can also perform incremental editing i.e., append or remove extra text lines, by using space character (`` ") on top of visible texts (C). The process can repeat until the end-user is satisfied with the final generated results (D).}}
    
    \label{fig:simulate_workflow}
\end{center}%
}]


\begin{abstract}
\vspace{-2mm}
    Textual image generation spans diverse fields like advertising, education, product packaging, social media, information visualization, and branding. Despite recent strides in language-guided image synthesis using diffusion models, current models excel in image generation but struggle with accurate text rendering and offer limited control over font attributes. In this paper, we aim to enhance the synthesis of high-quality images with precise text customization, thereby contributing to the advancement of image generation models. We call our proposed method \textit{CustomText}. Our implementation leverages a pre-trained TextDiffuser model to enable control over font color, background, and types. Additionally, to address the challenge of accurately rendering \textit{small sized fonts}, we train the ControlNet model for a consistency decoder, significantly enhancing text-generation performance. We assess the performance of CustomText in comparison to previous methods of textual image generation on publicly available CTW-1500 dataset and a self-curated dataset for small-text generation, showcasing superior results.
\end{abstract}
\begin{figure*}[t]
  \centering
   \includegraphics[width=0.75\textwidth]{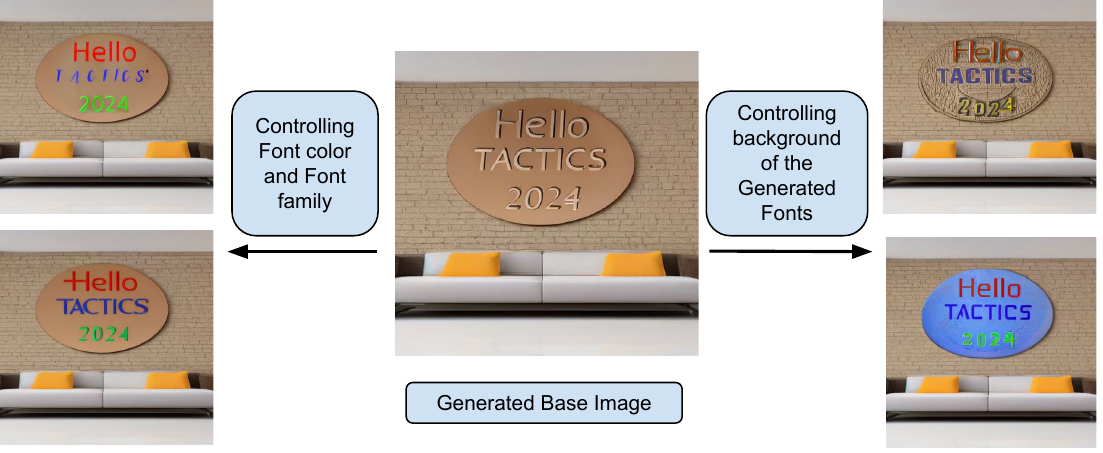}
  \caption{Example demonstrating the control of our proposed method CustomText over fonts color, fonts types and fonts background on the base image.}
  \label{fig:font_controls}
\end{figure*}

\vspace{-5mm}
\section{Introduction}
\label{sec:intro}

The domain of text-to-image synthesis has witnessed remarkable advancements, with diffusion models emerging as a key paradigm in this domain. These models provide  quality and diversity of generated content through the diffusion process. Diffusion models such as DALLE-2~\cite{ramesh2022hierarchical}, Imagen~\cite{saharia2022photorealistic}, and Stable Diffusion~\cite{rombach2022high}, leverage the semantic richness inherent in textual prompts. Recent breakthroughs in diffusion models 
~\cite{mou2023t2i},\cite{gal2022image},\cite{rombach2022high},\cite{ruiz2023dreambooth},\cite{couairon2022diffedit},\cite{hertz2022prompt} offer distinct editing control advantages.



Despite the significant progress in diffusion-based methods,  \textit{generating textual content within the image with complex font attributes} is still a challenge. The generation of textual-images (or images with textual information) has diverse applications in industries such as entertainment, advertising (see Figure.~\ref{fig:simulate_workflow}), education, and product packaging. Creating high-quality text-images in diverse formats such as posters, book covers, etc, conventionally requires professional skills and iterative design processes. 
 Traditional digital methods~\cite{adobephotoshop} involving manual labor may often yield unnatural appearances or digital artifacts due to complex background textures and lighting variations. Moreover, the direct rendering of text in a proper font on top of the generated image lacks harmonization of text with the background and does not produce a visually appealing image. 
 
\begin{figure*} [t]
  \begin{subfigure}[t]{.3\textwidth}
    \centering
    \includegraphics[width=1.0\textwidth]{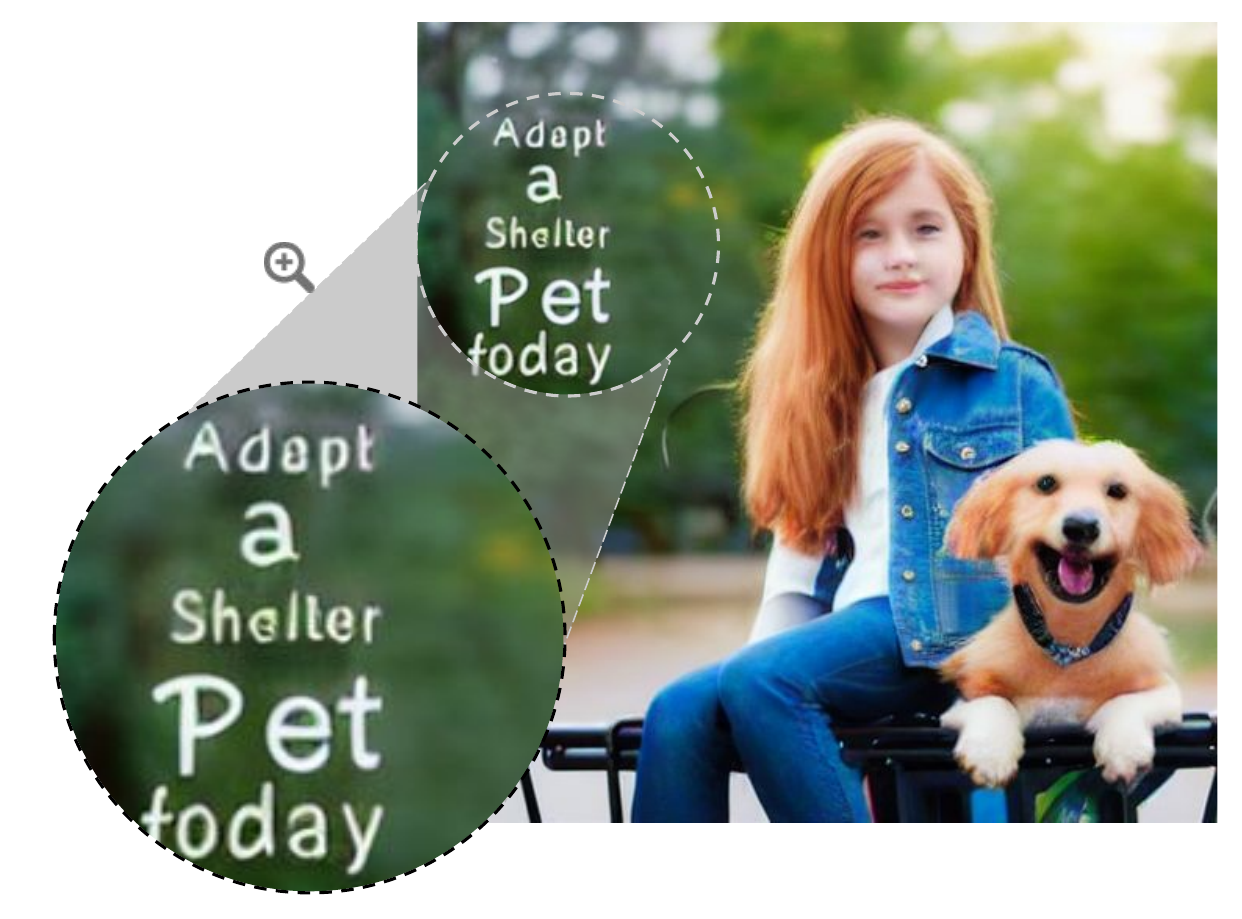}
    \caption{TextDiffuser decoder~\cite{chen2023textdiffuser}}
    \label{fig:comapre_a}  
  \end{subfigure}
  \hfill
  \begin{subfigure}[t]{.3\textwidth}
    \centering
    \includegraphics[width=1.0\textwidth]{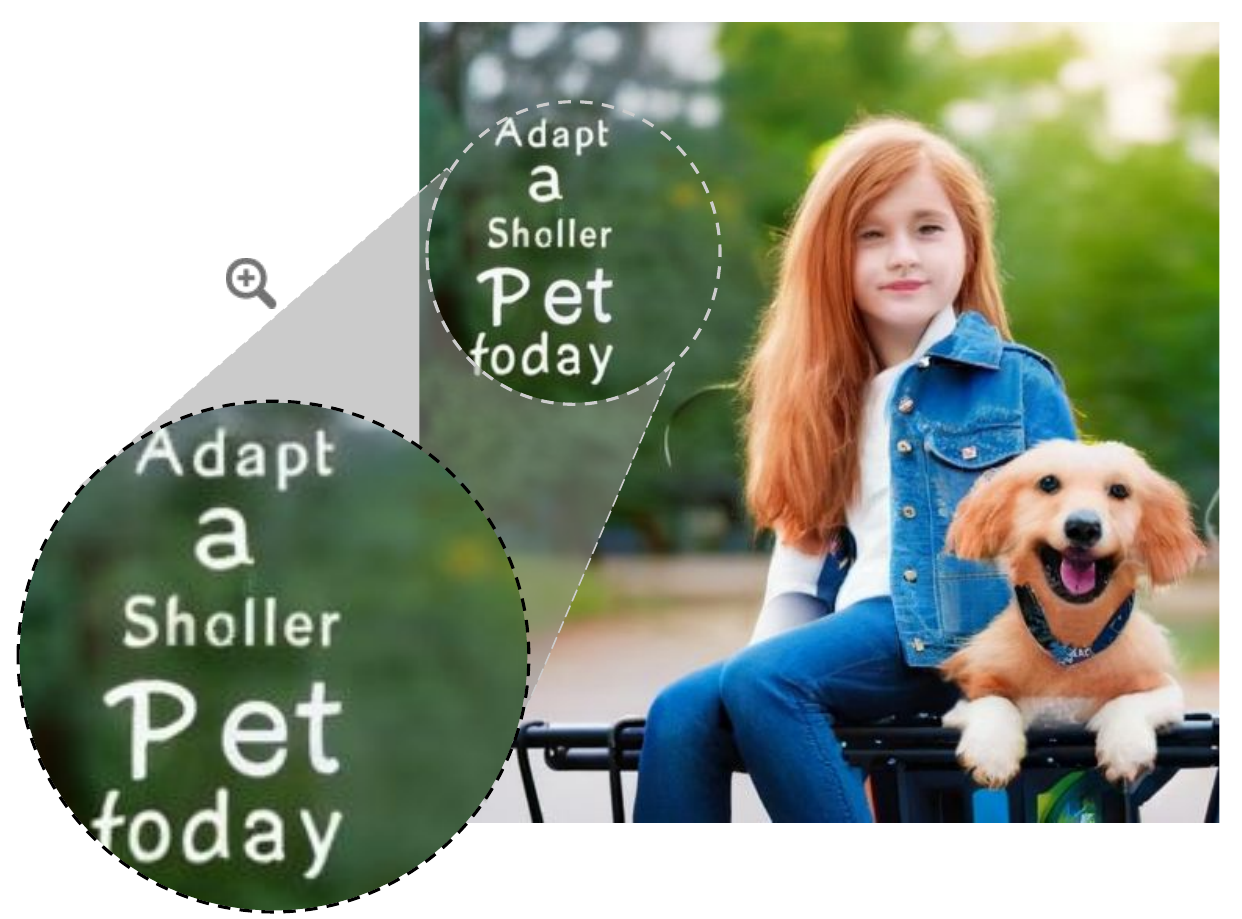}
    \caption{DALLE-3 Consistency decoder~\cite{song2023consistency}}
    \label{fig:comapre_b}
  \end{subfigure}
  \hfill
  \begin{subfigure}[t]{.3\textwidth}
    \centering
    \includegraphics[width=1.0\textwidth]{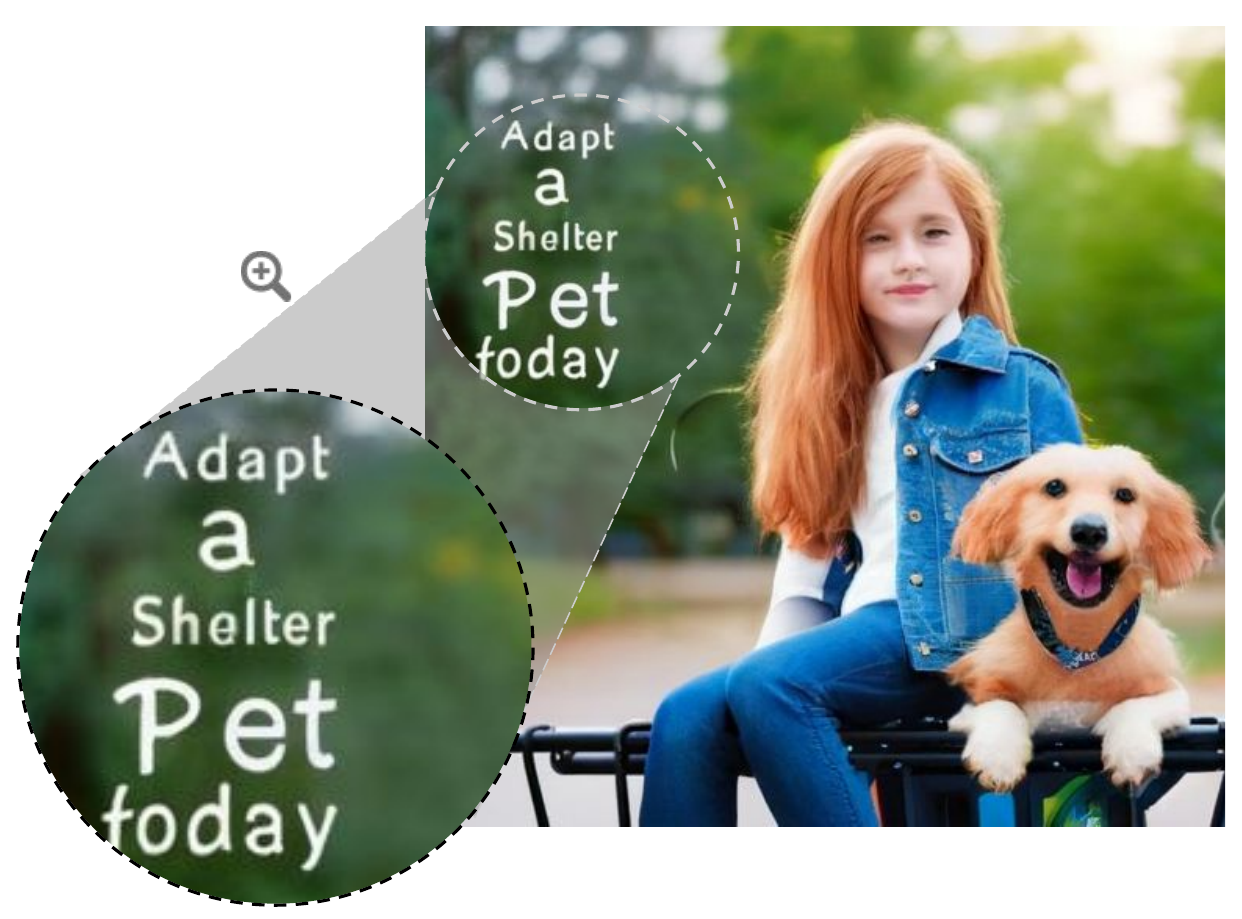}
    \caption{CustomText decoder}
    \label{fig:comapre_c}
  \end{subfigure}  
  \caption{An example use-case of textual-image generation such as an advertisement with text ``Adapt a Shelter Pet today". The images \ref{fig:comapre_a}, \ref{fig:comapre_b} and \ref{fig:comapre_c} are generated using TextDiffuser decoder, DALLE-3 Consistency decoder and our proposed CustomText decoder, respectively. The CustomText shows superior control over accurate text-generation as it is evident that it writes ``Shelter" accurately in comparison to other methods.}
  \label{fig:usecase-ad}  
\end{figure*}

 Current efforts to enhance text rendering quality have turned to diffusion models, exemplified by pioneering frameworks like Imagen~\cite{saharia2022photorealistic}, eDiff-I~\cite{balaji2022ediffi}, and DeepFloyd~\cite{deepfloyd}. These models have demonstrated improved text generation capabilities by using better text encoders, e.g., leveraging T5 series text encoders~\cite{raffel2020exploring} over the CLIP~\cite{radford2021learning} text encoder.
Additionally, better trained language models demonstrate increasing capability for text-rendering when they increase the parametric strength, e.g. Parti~\cite{yu2022scaling} shows a dramatic improvement in text-rendering when the model parameters increase from 350M to 200B. However, these models lack comprehensive control over the generation process. 
Some concurrent works, such as GlyphDraw~\cite{ma2023glyphdraw} and TextDiffuser~\cite{chen2023textdiffuser}, aim to enhance control by conditioning on the location and structures of Chinese characters and English characters, respectively. However, the limitation of not supporting multiple text bounding-box generation restricts the applicability of GlyphDraw~\cite{ma2023glyphdraw} to various text-image scenarios, such as posters and book covers. On the other hand, TextDiffuser~\cite{chen2023textdiffuser} addresses the challenges in creating multiple text boxes within images, but still fails in the generation of dense and small text (see Figure~\ref{fig:usecase-ad}(a)). Additionally, both of these models cannot control font attributes during generation.
In a very recent work by~\cite{liu2024glyph}, authors have proposed an architecture where a text encoder is trained over the synthetic dataset to generate small and dense texts. Though the results are very impressive, the model is very resource-intensive.

To address these challenges, we present a novel approach named CustomText for generating images with customized font attributes including font type, color, and background (refer to Section~\ref{sec:methodology}).
Our key contributions in this paper are as follows:
\begin{itemize}
    \item We propose a novel inference pipeline to \textit{control text attributes} like color, type, size, and background.
    \item We supplement  consistency model decoder~\cite{betker2023improving} with a  ControlNet-based architecture to provide better \textit{rendering of small-sized fonts} 
\end{itemize}

We assess the reconstruction performance of the trained decoder over the test-split of CTW-1500~\cite{yuliang2017detecting} dataset and show that it has improved performance over methods such as ControlNet-canny~\cite{zhang2023adding} and TextDiffuser~\cite{chen2023textdiffuser}.  We also demonstrate the efficacy of our proposed method for small-font text generation over the self-created \textit{SmallFontSize} dataset containing 200 examples of textual prompts with corresponding character maps. 

\begin{figure*}[!h]
  \centering
   \includegraphics[width=0.85\textwidth]{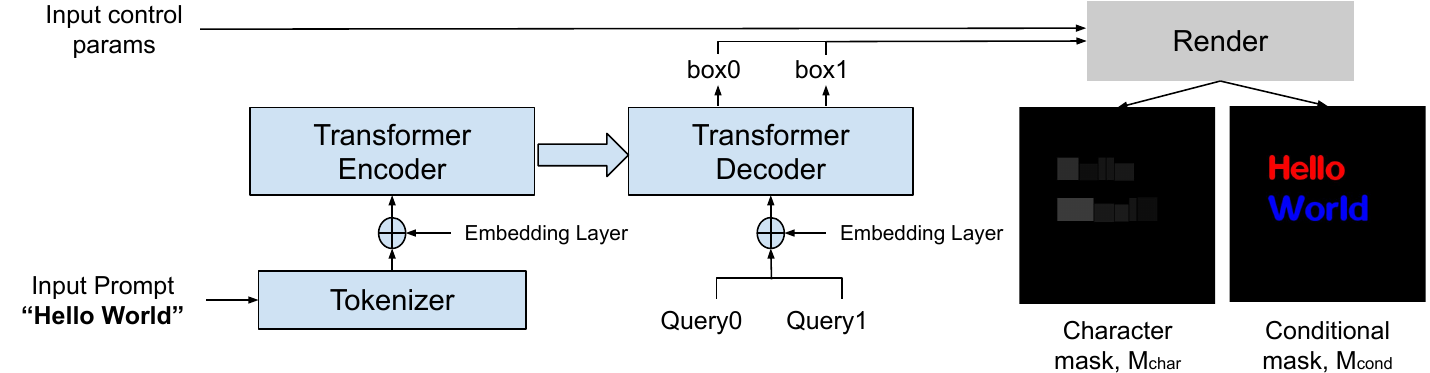}
  \caption{Stage 1 pipeline for the generation of character mask ($M_{char}$) and conditional mask ($M_{cond}$) using input textual prompt and control parameters defining font-attributes. The transformer encoder-decoder architecture takes input prompt and for each word, unique non-overlapping bounding box is extracted. The input control parameters define the different font-attributes such as color, type, background, which enable the renderer to generate the desired conditional mask.}
  \label{fig:stage1}
 \end{figure*}

\begin{figure*}[!h]
  \centering
   \includegraphics[width=0.8\textwidth]{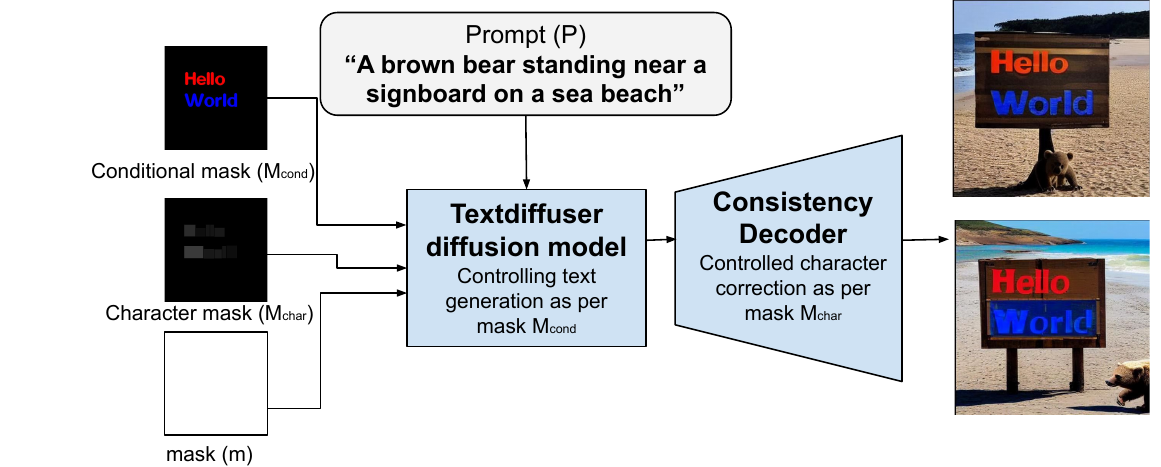}
  \caption{Stage 2 pipeline of our proposed CustomText method for generating images with the desired text attributes by using character mask ($M_{char}$), conditional mask ($M_{cond}$) and textual prompt (P). Please note that the white region in mask $m$ represents the region where the user wants to perform generation.}
  \label{fig:stage2}
 \end{figure*}

\section{Proposed Method: CustomText}
\label{sec:methodology}
We aim to generate images or perform image in-painting with controlled text generations. This control extends to font attributes such as type, size, color, and background, all of which need to seamlessly integrate into a given reference image layout (in the case of image in-painting). Our objective is to ensure a resulting image with a high degree of harmonization and photo-realism. To achieve this objective, we identify two main areas of improvement over current methods such as ~\cite{chen2023textdiffuser}, ~\cite{ma2023glyphdraw}: (1) Provide explicit control of text attributes during textual image generation. (2) For layouts that consist of small-sized characters, generate visually clear characters.



\subsection{Text Customization with Attribute Control}
Our task is to generate image $x$,  given a textual prompt $P$ with text-content $g$ and a mask $m$ representing a region of interest for $g$. This generated image should incorporate the formatted text $F(g)$ within the designated region, where $F$ represents a set of formatting functions containing control parameters that govern the manipulation of font color, type, and background. We propose a two-stage pipeline for custom text generation.



 
\subsubsection{Pipeline: First stage}
\label{sec:pipe-stage1}
In the first stage, as shown in Figure~\ref{fig:stage1}, we obtain two masks, namely, character mask $M_{char}$ and conditional mask $M_{cond}$. The character mask $M_{char}$ defines the spatial position for the generated text $g$, where we allocate a rectangular box for each character generation. The conditional mask $M_{cond}$ specifies the necessary attributes of $g$ based on the functions in $F$. This stage takes textual prompt $P$ as input, with generated text $g$ specified through single quotes. We utilize a Layout Transformer based architecture~\cite{gupta2021layouttransformer} to predict a bounding box $B$ for $g$ in the mask image. Subsequently, the character mask is created by using this bounding box information $M_{char} \leftarrow (B, g)$. This information is also utilized by the rendering module for each character and combined with the control parameters defined by $F$, to obtain the conditional mask, $M_{cond} \leftarrow (B, F(g))$.



\subsubsection{Pipeline: Second Stage}
\label{sec:pipe-stage2}
In the second stage, as shown in Figure~\ref{fig:stage2}, we utilize the pre-trained model of TextDiffuser ~\cite{chen2023textdiffuser}, which provides spatial control (only) using the character mask. We modify the inference pipeline to provide additional attribute control in the text generation process through conditional mask $M_{cond}$.
For the purpose of generation, the diffusion model is initialized with random Gaussian noise, $x_{T}$, and character mask, $M_{char}$. We utilize DDPM~\cite{ho2020denoising} scheduler to compute image inversion of conditional mask $M_{cond}$. Our aim is to infuse the effect (essence) of conditional mask features in the denoising steps of $x_{T}$. Hence, we obtain the noisy samples $M_{cond}$ for different timesteps $t$ in $[0, T]$ to obtain a list of latents, $Q_{cond}$.

To ensure the effects of $M_{cond}$ in the generation process, the self-attention map of $x_{t}$ must encapsulate the effects of $q_{cond,t}$, where $q_{cond,t}$ represents  $Q_{cond}$ sample at timestep $t$. 
To affect this, a progressively reducing time-step dependent weighted $q_{cond,t}$ is added to $x_{t}$ for different $t$.   This weighted addition diminishes the impact of $M_{cond}$  on the final stages of the generation and helps avoid sharp boundaries between the rendered image and the conditional image. 

\begin{equation}
    w_{c,t} = W(M_{char}, t)
\end{equation}
\begin{equation}
    x_{t}^{'} = x_{t} * (1-w_{c,t}) + q_{cond, t} * w_{c,t}
\end{equation}

Thus, a weighted forward-propagated conditional image $q_{cond,t}$ from $Q_{cond}$ at time-step $t$ is introduced into the image latent $x_{t}$ to control character properties, including font types, color, and background. A carefully designed weighting function alleviates abrupt transitions on the text boundaries.

\subsection{Generation of accurate small-sized text}

Many latent diffusion models employ Variational Autoencoder (VAE)~\cite{kingma2013auto} network to transform images into low-dimensional latent spaces, to enhance computational efficiency during the training and inference of diffusion models. However, when images are compressed into lower-dimensional latent spaces, the fine details might be lost. This can result in the generated images not accurately reproducing the original data, which could be problematic where precise details, such as the rendering of small-sized fonts, are crucial. To address this issue, we propose two alternate approaches, where one approach uses a VAE-based latent decoder with split and merge on latents, and another approach uses a consistency decoder model coupled with control nets.


\begin{figure}[!h]
  \centering
   \includegraphics[width=0.4\textwidth]{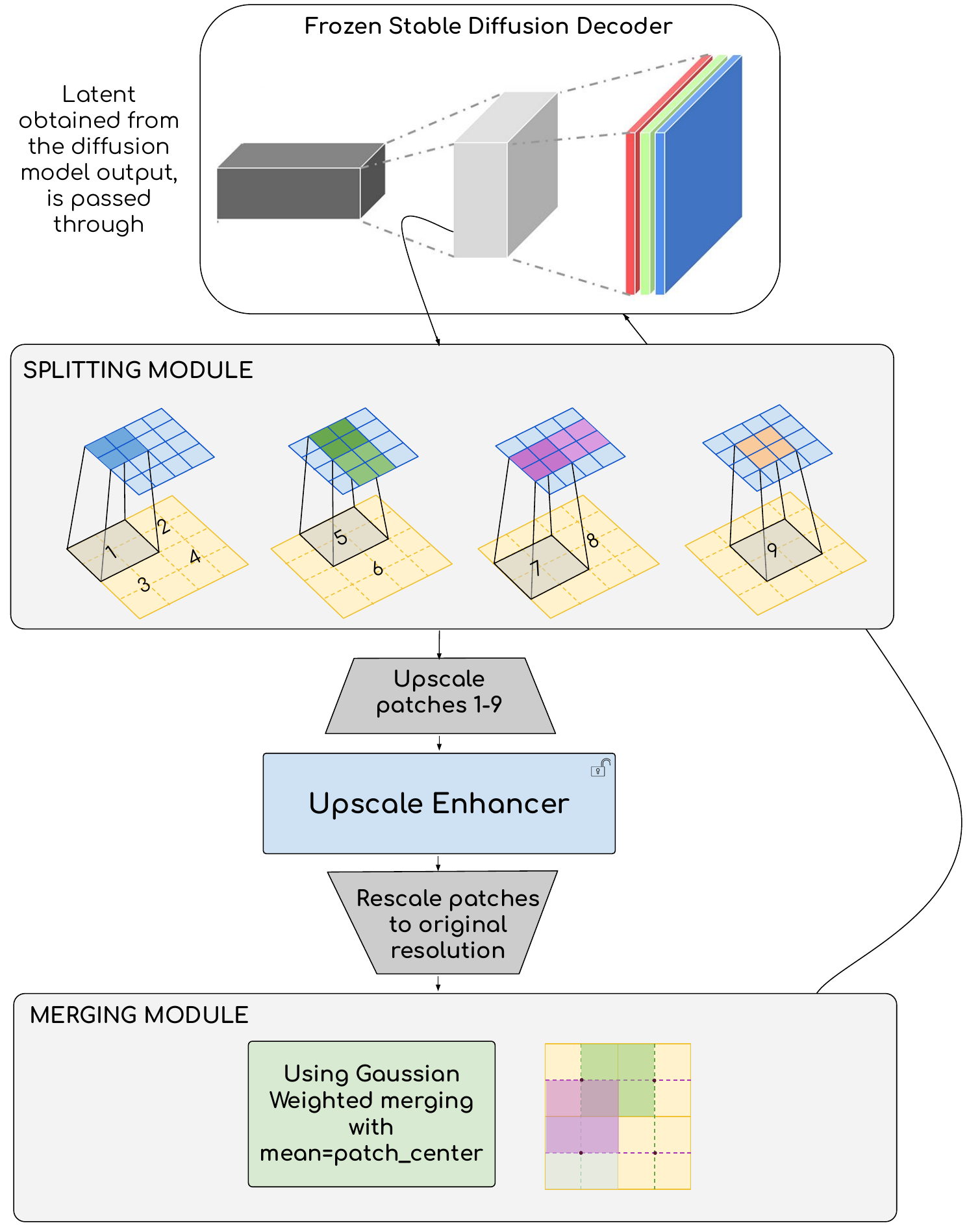}
  \caption{\small{VAE based Decoder Enhance model. The latent coming from the diffusion model is splitted into $9$ overlapping fragments in the Splitting Module. The resized fragments are upscaled and a trainable model learns to enhance the quality of the generation. Finally, the enhanced fragments are assembled to create final image in the original resolution by the merging module.}}
  \label{fig:decoder_enhancing}
 \end{figure}

 \begin{figure}[!h]
  \centering
   \includegraphics[width=0.4\textwidth]{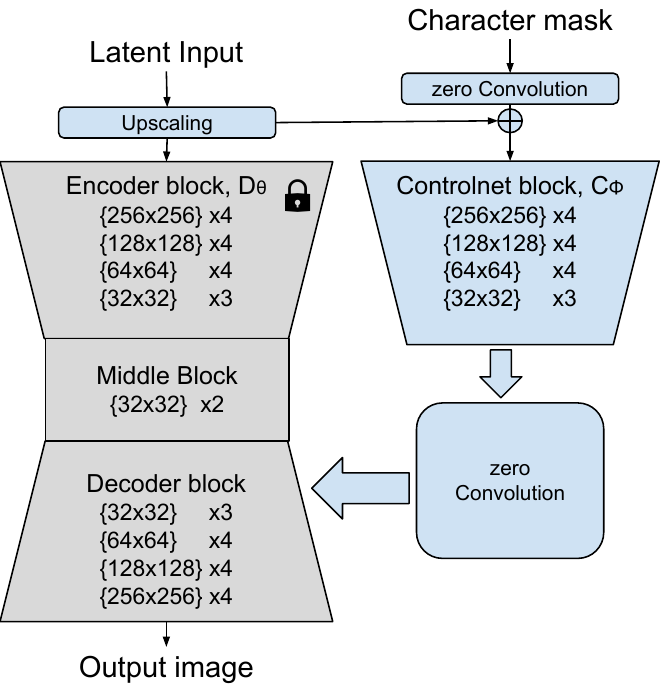}
  \caption{\small{Modified Consistency Decoder architecture. The weights of original decoder $D_{\theta}$ are locked and a trainable ControlNet block $C_{\phi}$ is appended.}}
  \label{fig:controlnet_arch}
 \end{figure}

\subsubsection{VAE based Decoder Enhance model}

The intuition of our approach comes from the observation that the regions with small-sized characters are often subject to rendering anomalies. These small-sized characters can be treated as large-sized characters if they are scaled individually. The cropped and scaled image patch may have upscale blurs because of low resolution. However, if we can fix the blurring of the upscaled patch and consequently downscale and append it back to the original resolution, then reconstruction quality should improve. Taking cues from this intuition, we propose an architecture, as shown in Figure~\ref{fig:decoder_enhancing}, consisting of three parts: Splitting Module, Upscale Enhancer, and Merging module, which takes the latent feature map as an input and produces the reconstructed image.



\textbf {\textit{Splitting Module}}: In this module, there are no learnable parameters. It takes the input of feature map $D(x_{g}) \in \mathbb{R}^{(h^{'}\times w^{'})\times d}$, where $D$ is the VAE decoder from a trained stable diffusion model and $h^{'} \times w^{'}$ is the
size of the input feature map. The splitting module crops the feature map into $9$ overlapping patches, such that each patch's dimension is $h^{'}/2 \times w^{'}/2$. These patches are then upscaled using bilinear interpolation back to their original image resolution.

\textbf {\textit{Upscale Enhancer Module}}: Here, the objective is to refine the upscaled image. To achieve this, we use the sequence of zero convolution operations. The `zero convolution' refers to a $1 \times 1$ convolution with both weight and bias initialized as zeros. The choice of 0-convolution is deliberate because the initialization of weights and biases with $0$ will not alter the input feature map in the forward pass and, hence, it does not distort the original pre-trained VAE's decoder output. However, after one iteration update with non-zero reconstruction loss, weight and bias terms are updated to non-zero, and the normal training procedure continues. Thus, zero-convolution provides a safer way to enhance reconstruction without modifying the original weight of decoders. As inspired from~\cite{chen2023textdiffuser}, the objective function used to train the model consists of reconstruction loss (L2 loss), along with character-aware loss.

\textbf {\textit{Merging Module}}: Finally, the upscaled enhanced patches should be merged to the final resolution. Since the patch resolution matches the final resolution, merging the patches to their original resolution (one used during splitting) will degrade the enhancements. Thus the patches are merged, and weighted-addition aggregates the enhancements with the decoder's final output. To merge the overlapping patches, we use weighted merging to carefully allot more weight to the patches that are at the center of the attending pixels in the final image.


While this method is effective in enhancing the reconstruction quality and performs better than the vanilla decoder, it still requires a large training dataset to learn. Furthermore, another drawback of the method is that the splitting and merging dimensions of patches are pre-defined, and therefore they might not always overlap with actual small character regions.

\begin{figure*}[t]
  \centering
   \includegraphics[width=1.0\textwidth]{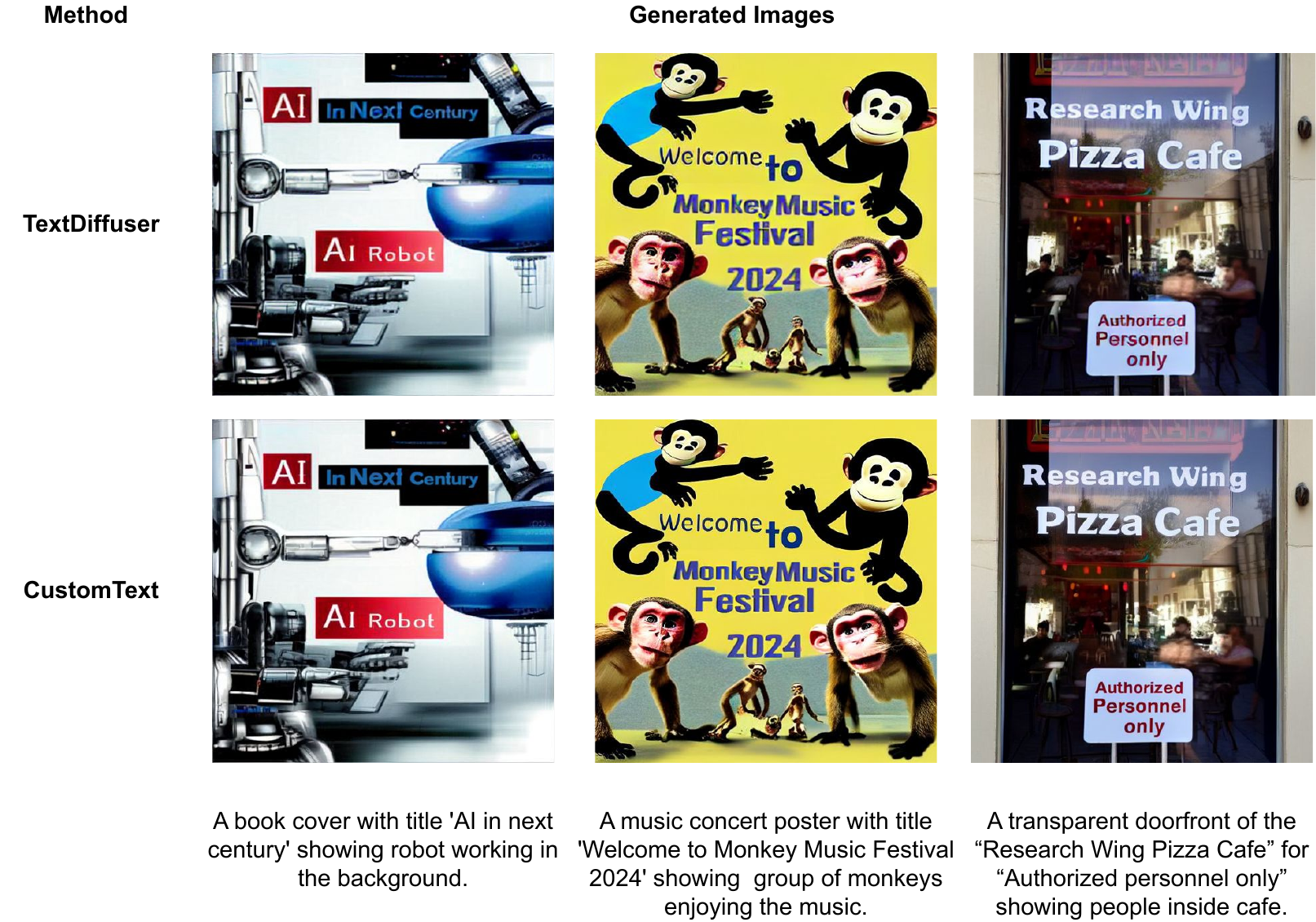}
  \caption{Qualitative comparison of text generation quality with TextDiffuser~\cite{chen2023textdiffuser} and CustomText (ours).}
  \label{fig:qualitative_comparison}
 \end{figure*}

\begin{table*}[t]
\centering
\begin{tabular}{|c|ccc|}
\hline
\textbf{}  &
  \multicolumn{3}{c|}{\textbf{CTW-1500}} \\
  \hline  
\textbf{Method}  &
  \multicolumn{1}{c|}{\textbf{MSE $\downarrow$}} &
  \multicolumn{1}{c|}{\textbf{PSNR $\uparrow$}} &
  \multicolumn{1}{c|}{\textbf{SSIM $\uparrow$}} \\ \hline
\textbf{Controlnet-canny~\cite{zhang2023adding}} &
  \multicolumn{1}{c|}{0.033} &
  \multicolumn{1}{c|}{17.82} &
  \multicolumn{1}{c|}{0.656} \\   
\textbf{TextDiffuser~\cite{chen2023textdiffuser}} &
  \multicolumn{1}{c|}{0.031} &
  \multicolumn{1}{c|}{17.82} &
  \multicolumn{1}{c|}{0.6601} \\ 
  \hline
\textbf{VAE Decoder Enhance (ours)} &
  \multicolumn{1}{c|}{0.027} &
  \multicolumn{1}{c|}{18.17} &
  \multicolumn{1}{c|}{0.6874} \\ 
\textbf{CustomText (ours)} &
  \multicolumn{1}{c|}{\textbf{0.019}} &
  \multicolumn{1}{c|}{\textbf{21.33}} &
  \multicolumn{1}{c|}{\textbf{0.712}} \\   
\hline
\end{tabular}
\caption{Quantitative comparison of Reconstruction performance on CTW-1500 test set.}
\label{table:enhancer_result}
\end{table*}

\begin{table*}[t]
\centering
\begin{tabular}{|c|ccc|}
\hline
\textbf{}  &
  \multicolumn{3}{c|}{\textbf{CTW-1500}} \\
  \hline  
\textbf{Method}  &
  \multicolumn{1}{c|}{\textbf{Precision}} &
  \multicolumn{1}{c|}{\textbf{Recall}} &
  \multicolumn{1}{c|}{\textbf{F1}} \\ \hline
\textbf{Controlnet-canny~\cite{zhang2023adding}} &
  \multicolumn{1}{c|}{0.7355} &
  \multicolumn{1}{c|}{0.7581} &
  \multicolumn{1}{c|}{0.7466} \\   
\textbf{TextDiffuser~\cite{chen2023textdiffuser}} &
  \multicolumn{1}{c|}{0.7355} &
  \multicolumn{1}{c|}{0.7581} &
  \multicolumn{1}{c|}{0.7466} \\ 
  \hline
\textbf{VAE Decoder Enhance (ours)} &
  \multicolumn{1}{c|}{0.7355} &
  \multicolumn{1}{c|}{0.7581} &
  \multicolumn{1}{c|}{0.7466} \\ 
\textbf{CustomText (ours)} &
  \multicolumn{1}{c|}{\textbf{0.748}} &
  \multicolumn{1}{c|}{\textbf{0.762}} &
  \multicolumn{1}{c|}{\textbf{0.7549}} \\   
\hline
\end{tabular}
\caption{Quantitative comparison of OCR performance using EasyOCR over reconstructed image.}
\label{table:enhancer_result2}
\end{table*}

\begin{table*}[t]
\centering
\begin{tabular}{|c|cccc|}
\hline
\textbf{}  &
  \multicolumn{4}{c|}{\textbf{SmallFontSize Dataset}} \\
  \hline 
\textbf{Method}  &
  \multicolumn{1}{c|}{\textbf{Precision}} &
  \multicolumn{1}{c|}{\textbf{Recall}} &
  \multicolumn{1}{c|}{\textbf{F1}} &
  \multicolumn{1}{c|}{\textbf{ClipScore}}
  \\ \hline
\textbf{StableDiffusion~\cite{rombach2022high}} &
  \multicolumn{1}{c|}{0.0936} &
  \multicolumn{1}{c|}{0.1174} &
  \multicolumn{1}{c|}{0.1041} &
  \multicolumn{1}{c|}{0.512} \\
\textbf{Controlnet-canny~\cite{zhang2023adding}} &
  \multicolumn{1}{c|}{0.6332} &
  \multicolumn{1}{c|}{0.6572} &
  \multicolumn{1}{c|}{0.645} &
  \multicolumn{1}{c|}{0.6321} \\
\textbf{TextDiffuser~\cite{chen2023textdiffuser}} &
  \multicolumn{1}{c|}{0.792} &
  \multicolumn{1}{c|}{0.7863} &
  \multicolumn{1}{c|}{0.7891} &
  \multicolumn{1}{c|}{\textbf{0.6407}} \\
\hline
\textbf{VAE Decoder Enhance (ours)} &
  \multicolumn{1}{c|}{0.7894} &
  \multicolumn{1}{c|}{0.7911} &
  \multicolumn{1}{c|}{0.7902} &
  \multicolumn{1}{c|}{\textbf{0.6407}} \\
\textbf{CustomText (ours)} &
  \multicolumn{1}{c|}{\textbf{0.8131}} &
  \multicolumn{1}{c|}{\textbf{0.815}} &
  \multicolumn{1}{c|}{\textbf{0.814}} &
  \multicolumn{1}{c|}{0.6392} \\
\hline
\end{tabular}
\caption{Quantitative comparison of OCR performance using EasyOCR over text-regions of image over SmallFontSize Dataset.}
\label{table:enhancer_texts_result_smalldataset}
\end{table*}

\subsubsection{Consistency Decoder with Character Map Guidance}

We propose an alternative approach to address the small-sized text generation, where we use a decoder that can generate high-quality small characters from the latent representations. Our approach, as shown in Figure~\ref{fig:controlnet_arch}, involves introducing a \emph{Character Mask} to a pre-trained Consistency decoder Model (CM)~\cite{betker2023improving}\cite{song2023consistency} that utilizes the semantic information of the characters. The intuition is that by incorporating the character guidance map (initially utilized for text generation) into the ControlNet~\cite{zhang2023adding} architecture, the decoder gains additional guidance for effective character reconstruction. The CM decoder then proves effective in preserving the identity and style of input characters, generating realistic and diverse small characters within the latent space.

The consistency diffusion model consists of a decoder network $f_{\theta}$ that takes as input a noise tensor $z_{t}$ sampled from a Gaussian distribution $N(0, I)$, and outputs an image $x_{0}$ that corresponds to the starting point of the diffusion path trajectory. The model can generate images in one step by sampling a noise tensor $z_{T}$ from the final timestep of the diffusion process and passing it through the decoder network $f_{\theta}$. 

Given the pre-trained consistency model decoder $D_{\theta}(.,.)$, we freeze the parameters $\theta$ and introduce ControlNet model $C_{\phi}$, with trainable parameters $\phi$, as shown in the Figure~\ref{fig:controlnet_arch}. The architecture takes latent vector, $l_{t}$ as input for $D_{\theta}$ and character mask $M$ (as defined in Section~\ref{sec:pipe-stage1}) for $C_{\phi}$. By adding the ControlNet architecture, we define a new consistency model $D_{\{\theta , \phi\}}(.,.,.)$ and train it with the loss\cite{song2023consistency} \cite{song2023improved}, as defined in Eq.\ref{cons_loss}, which ensures stable training.

\begin{multline}
\label{cons_loss}
    L_{C}(\phi) = E_{l, l_{t}, M, n}[\lambda(t_n)d(D_{\{\theta , \phi\}}(l_{t_{n+1}}, t_{n+1}, M), \\
    D_{\{\theta , \phi\}^-}(l_{t_{n}}, t_n, M))]
\end{multline}
Here, $E[.]$ denotes the expectation over all random variables and $d(x, y)$ is $l_{2}$ squared distance. ${\{\theta,\phi\}^-}$ $\leftarrow$ stopgrad(${\{\theta, \phi\}}$) and only $\phi$ is kept trainable in the process. Thus, the trained architecture which takes the character mask as a guide helps to generate better quality of small sized characters.

\section{\uppercase{Experiments \& Results}}
\label{sec:results}

\begin{figure*} [ptbh]
  \hspace{10mm}
  \begin{subfigure}[t]{.4\textwidth}
    \centering
    \frame{\includegraphics[width=8.0cm,height=4cm]{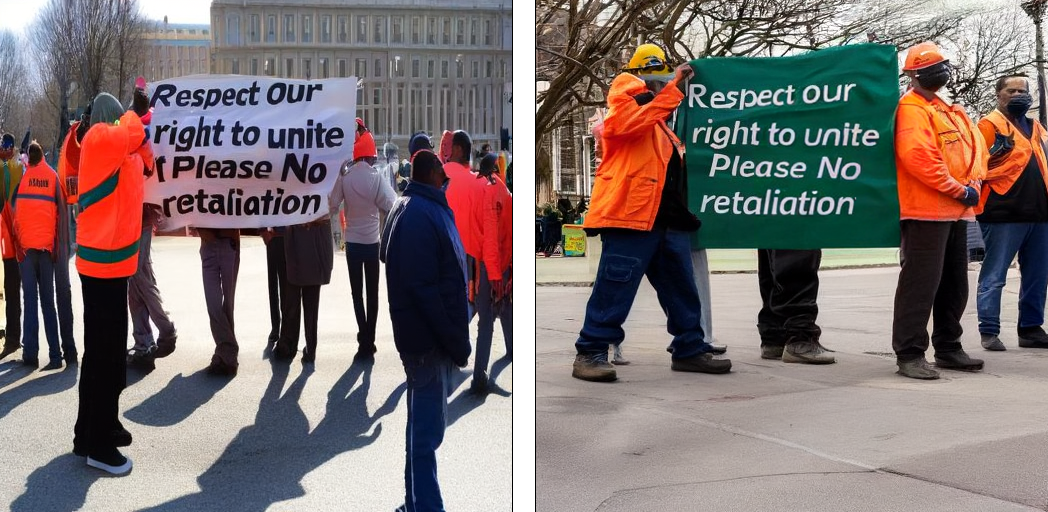}}    
    \caption{\small{Industrial workers in orange jacket protesting in front of green park with a signboard which says `\textit{Respect our right to unite Please No retaliation}' on a clear sunny day.}}
  \end{subfigure}
  \hfill
  \begin{subfigure}[t]{.4\textwidth}
    \centering
    \frame{\includegraphics[width=8.0cm,height=4cm]{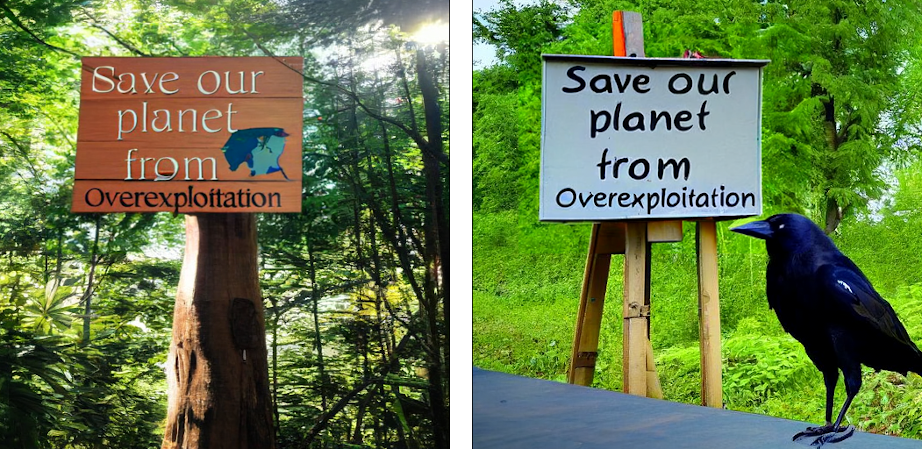}}
    \caption{\small{A wooden signboard saying `\textit{Save our planet from Overexploitation}' in the middle of jungle. A bird is sitting near the signboard.}}
  \end{subfigure}
  
  \medskip
 \hspace{10mm}
  \begin{subfigure}[t]{.4\textwidth}
    \centering
    \frame{\includegraphics[width=8.0cm,height=4cm]{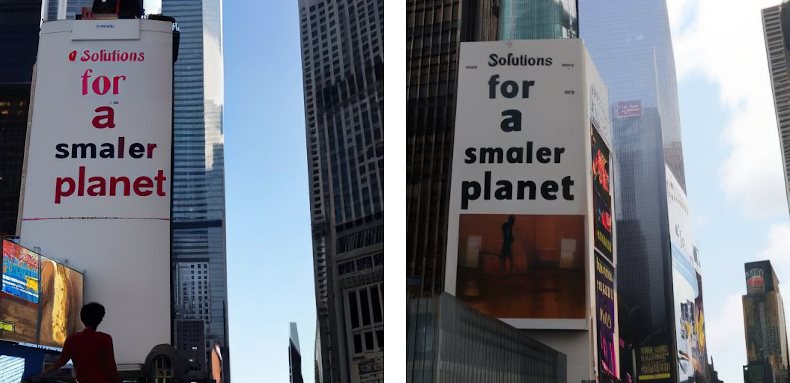}}    
    \caption{\small{A large advertisement board on a tall building in a metropolis saying `\textit{Solutions for a smaller planet}' with clear sunny day in the background.}}
  \end{subfigure}
  \hfill
  \begin{subfigure}[t]{.4\textwidth}
    \centering
    \frame{\includegraphics[width=8.0cm,height=4cm]{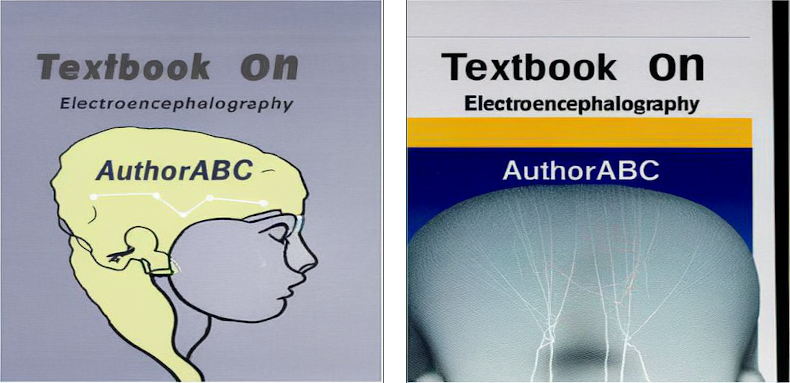}}
    \caption{\small{A cover of school text book with title `\textit{Textbook on Electroencephalography AuthorABC}'.}}
  \end{subfigure}

  \medskip
 \hspace{10mm}
  \begin{subfigure}[t]{.4\textwidth}
    \centering
    \frame{\includegraphics[width=8.0cm,height=4cm]{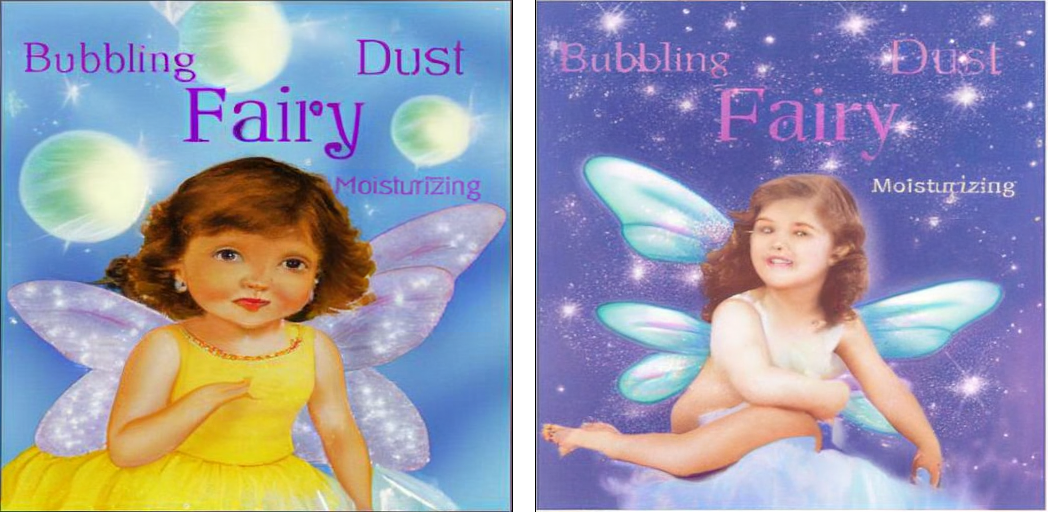}}   
    \caption{\small{A cover of children story book with title `\textit{Bubbling Fairy Dust Moisturizing}'.}}
  \end{subfigure}
  \hfill
  \begin{subfigure}[t]{.4\textwidth}
    \centering
    \frame{\includegraphics[width=8.0cm,height=4cm]{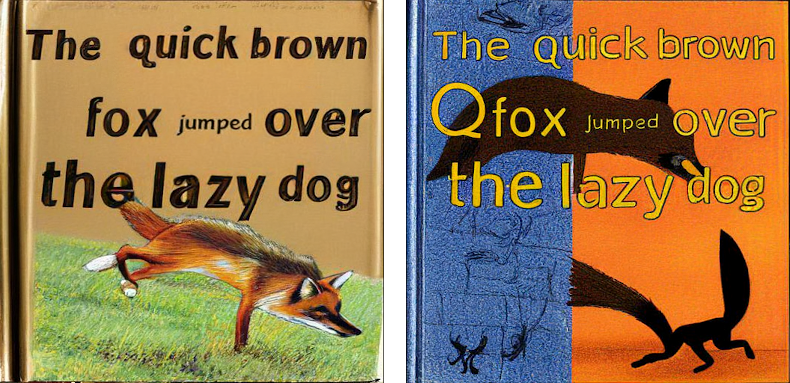}}
    \caption{\small{A cover of children story book with title `\textit{The quick brown fox jumped over the lazy dog}'.}}
  \end{subfigure}

  \medskip
  \hspace{10mm}
  \begin{subfigure}[t]{.4\textwidth}
    \centering
    \frame{\includegraphics[width=8.0cm,height=4.2cm]{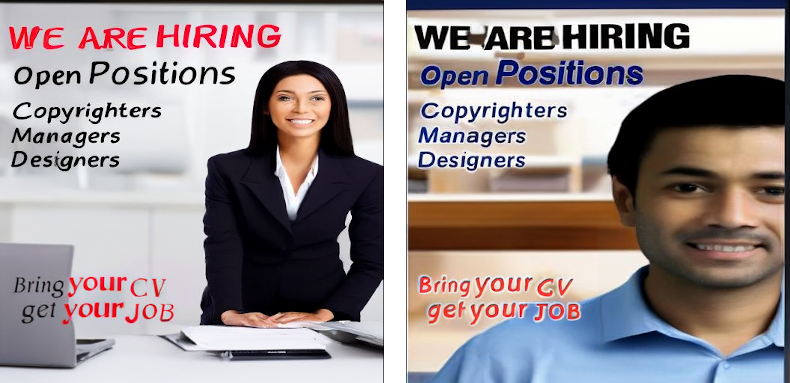}}    
    \caption{\small{A job vacancy advertisement. In the advertisement, `\textit{WE ARE HIRING Open Positions Copyrighters Managers Designers Bring your CV get your Job}' are printed. An employee working in the office is visible in the ad.}}
  \end{subfigure}
  \hfill
  \begin{subfigure}[t]{.4\textwidth}
    \centering
    \frame{\includegraphics[width=8.0cm,height=4.2cm]{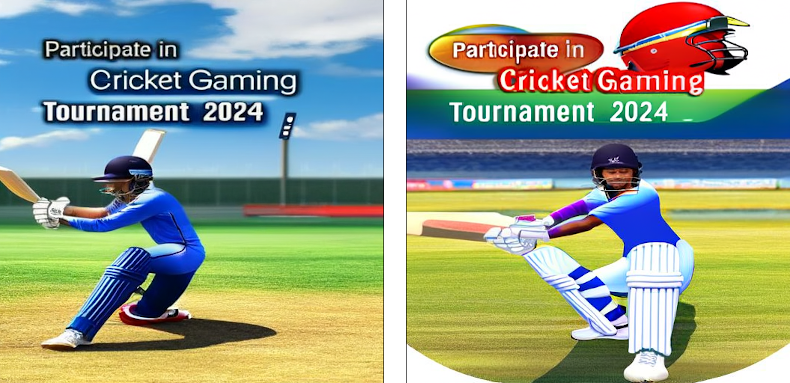}}
    \caption{\small{An advertisement for cricket game tournament which says `\textit{Participate in Cricket Gaming Tournament 2024}'. An image of batsman wearing helmet in blue jersey is shown in batting stance.}}
  \end{subfigure}  
 
  \caption{Qualitative results of CustomText (ours) for generating textual images containing varied-sized fonts over SmallFontSize Dataset.}
  \label{fig:plot}
\end{figure*}

\subsection{Dataset}
We have used the publicly available train and test split of CTW-1500~\cite{yuliang2017detecting} dataset to train the consistency decoder and evaluate the performance of CustomText, respectively. Additionally, we created a custom dataset namely SmallFontSize dataset to showcase the effectiveness of our method on generating small-sized fonts. The dataset contains 200 examples of textual prompts for generating small-sized texts in the images along with spatial character maps.

\subsection{Implementation}
We use the pre-trained TextDiffuser model and stable-diffusion-v1-5 model for our pipeline. For the decoder, we use the pre-trained DALLE-3 consistency decoder available publicly. Since, the original consistency decoder often distorts the small-sized characters in the decoded image, we tried to provide character map assistance for its correction (Refer Figure~\ref{fig:controlnet_arch}). We have followed original ControlNet~\cite{zhang2023adding} architecture to modify the consistency model. We have trained the model for 3500 steps over CTW1500 dataset, with effective batch size of 96, using gradient accumulation. During inference, the generated image resolution of 512$\times$512 is used, which is computed over CFG scale $7.5$.

\subsection{Results}
To evaluate the reconstruction performance of the trained model, we have used MSE~\cite{mse_wiki}, PSNR~\cite{psnr_wiki} and SSIM~\cite{ssim_wiki} as metrics and evaluated over CTW-1500 test set. 
We present the comparison results of the image generation with the Controlnet-canny model, Textdiffuser and VAE based Decoder Enhance in Table~\ref{table:enhancer_result}. It is evident that our CustomText decoder performs best in all the three metrics. Furthermore, to evaluate readability, we have used EasyOCR~\cite{baek2019character}\cite{shi2016end} to verify the quality of the reconstructed image in Table~\ref{table:enhancer_result2}. Here, exact match of individual words is considered to compute the results. Additionally, we present the comparison results of CustomText for generating small-sized texts in the images in Table~\ref{table:enhancer_texts_result_smalldataset} over SmallFontSize dataset using OCR performance and ClipScore~\cite{hessel2022clipscore}. Although our ControlNet based Consistency decoder model outperforms other existing methods in the OCR results, however, original TextDiffuser model performs better in terms of Clipscore by a small margin of $0.0015$. If we compare between the CustomText decoder and Decoder enhance model, it's important to note that the CustomText decoder model has substantially fewer parameters. Thus, we expect to see a boost in performance of CustomText when trained on larger dataset with computationally heavier decoder enhance having more parameters. Due to the limited dataset constraint, we leave this experiments for future work.
Next, please note that we only have character map in the SmallFontSize dataset which guides the model to generate images, and thus, we could not compare on FID metric~\cite{fid}, as it requires ground truth images to compute/compare gaussian distributions. Furthermore, we qualitatively compare the performance of our proposed CustomText decoder against TextDiffuser~\cite{chen2023textdiffuser} decoder and DALLE-3 Consistency decoder~\cite{song2023consistency} for small text generation. Figure~\ref{fig:usecase-ad} compares the three generated images containing texts with varying font-sizes. When dealing with smaller characters, digital artifacts become more apparent, highlighting any spelling errors. However, not such large observation is visible in case of large size characters.

Additionally, we demonstrate the controlling properties of the CustomText over generated texts in Figure~\ref{fig:font_controls}. We also provide qualitative comparison results of CustomText with original TextDiffuser in Figure~\ref{fig:qualitative_comparison}. Few more examples of image generations using our CustomText are presented in Figure~\ref{fig:plot}.
Moreover, we demonstrate the application of our proposed CustomText method for automatically generating advertisements. In fields of digital marketing, the demand for visually compelling advertisements has increased. Thus, to meet this demand efficiently, its important to automate and simplify the creative design process. One example workflow for generating ads automatically is shown in Figure~\ref{fig:simulate_workflow}.

\section{Conclusion}
\label{sec:conclusion}
In digital marketing, there is an increased demand for compelling advertisements. Text messages are an integral part of most advertisements and marketing campaigns. While the generic diffusion-based generative models are becoming proficient in better text rendering, there is no inherent control over text placement and attributes. In this paper, we presented \textit{CustomText} as a method for generating images with high-quality customized fonts. Our approach provides novel architectures to solve this problem. Even as we feel that more research is required for fuller control through diffusion models, our experimental results highlight the superior attribute control of our method in the generation of text compared to the previous methods. We also demonstrated that our method supports \textit{incremental editing}, as shown in Figure~\ref{fig:simulate_workflow}, to obtain customized and better-quality textual images in content creation tasks such as posters, advertisements, and marketing materials. Going forward, we aim to provide support for multi-lingual characters as the current system is only trained for Latin alphabets.

{
\bibliographystyle{ieee_fullname}
\bibliography{PaperForReview}
}

\end{document}